\documentclass[journal]{IEEEtran}

\usepackage{graphicx}
\usepackage{amssymb}
\usepackage{caption}
\usepackage{subcaption}

\begin{document}

\title{One-Class Convolutional Neural Network}

\author{Poojan Oza,~\IEEEmembership{Student Member,~IEEE,}
        and Vishal M. Patel,~\IEEEmembership{Senior Member,~IEEE}
\thanks{P. Oza and V. M. Patel are with the Department of Electrical and Computer Engineering, Johns Hopkins University, Baltimore, MD, USA. Email: \{poza2, vpatel36\}@jhu.edu.
	 This work was supported by the NSF grant 1801435.}
}

\maketitle

\begin{abstract}
We present a novel Convolutional Neural Network (CNN) based approach for one class classification.  The idea is to use a zero centered Gaussian noise in the latent space as the pseudo-negative class and train the network using the cross-entropy loss to learn a good representation as well as the decision boundary for the given class.   A key feature of the proposed approach is that any pre-trained CNN can be used as the base network for one class classification.  The proposed One Class CNN (OC-CNN) is evaluated on the UMDAA-02 Face, Abnormality-1001, FounderType-200 datasets. These datasets are related to a variety of one class application problems such as user authentication, abnormality detection and novelty detection. Extensive experiments demonstrate that the proposed method achieves significant improvements over the recent state-of-the-art methods. The source code is available at : github.com/otkupjnoz/oc-cnn.
\end{abstract}

\begin{IEEEkeywords}
One Class Classification, Convolutional Neural Networks, Representation Learning.
\end{IEEEkeywords}

%
\IEEEpeerreviewmaketitle

\section{Introduction}
Multi-class classification entails classifying an unknown object sample into one of many pre-defined object categories.  In contrast, in one-class classification, the objective is to identify objects of a particular class (also known as positive class data or target class data) among all possible objects by learning a classifier from a training set consisting of only the target class data.  The absence of data from the negative class(es) makes the one-class classification problem difficult.  

One-class classification has  many applications such as anomaly or abnormality detection \cite{chandola2009anomaly}, \cite{akoglu2015graph}, \cite{sindagi2017domain}, \cite{you2017provable}, \cite{sabokrou2018adversarially}, novelty detection \cite{abati2018and}, \cite{pimentel2014review}, \cite{markou2003novelty}, and user authentication \cite{fathy2015face}, \cite{perera2016quickest}, \cite{guo2014face}, \cite{antal2015evaluation}, \cite{VMP_AA_SPM}, \cite{zperera2018efficient}.
For example, in novelty detection, it is normally assumed that one does not have \textit{a priori} knowledge of the novel class data.  Hence, the learning process involves only the target class data.

\begin{figure}
\centering
\includegraphics[height=3.0cm]{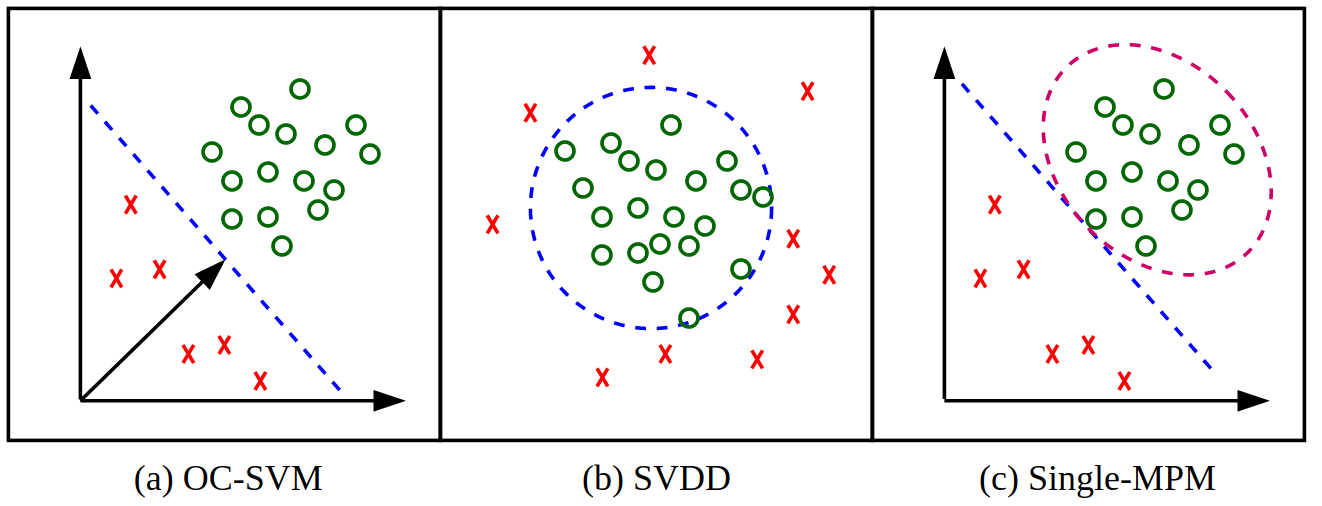}
\vskip -5.0pt \caption{A graphical illustration of popular statistical one-class classification methods. Green circles show the target class data, red crosses show the unknown data (i.e. anomaly, novelty, outlier etc.), blue doted lines/circles show the decision boundaries captured by the respective methods. Pink dotted line in Fig.~\ref{fig:other_methods}(c) shows the boundary of zero error of probability.  (a) OC-SVM, maximizing the margin of a hyperplane with respect to the origin.  (b) SVDD, finding a hypersphere that encloses the given data. (c) MPM, finding a hyperplane that minimizes the misclassification probability.}
\label{fig:other_methods}
\end{figure}

Various methods have been proposed in the literature for one-class classification.  In particular, many one-class classification methods are based on the Support Vector Machines (SVM) formulation \cite{scholkopf2000support}, \cite{markou2003novelty}, \cite{erfani2016high}.  SVMs are based on the concept of finding a boundary that maximizes the margin between two classes and are shown to work well for binary and multi-class classification.
However, in one-class problems the infromation regarding the negative class data is unavailable.  To deal with this issue, Scholkopf et al. \cite{scholkopf2001estimating} proposed one-class SVM (OC-SVM), which tackles the absence of negative class data by maximizing the boundary with respect to the origin.  Another popular approach inspired by the SVM formulation is Support Vector Data Description (SVDD) introduced by Tax et al. \cite{tax2004support}, in which a hypersphere that encloses the target class data is sought. Various extensions of OC-SVM and SVDD have been proposed in the literature over the years.  We refer readers to  \cite{khan2014one} for a survey of different one-class classification methods. Another approach for one-class classification is based on the Minimax Probability Machines (MPM) formulation \cite{lanckriet2002minimax}. Single class MPM \cite{ghaoui2003robust}, \cite{perera2018dual} seeks to find a hyper-plane similar to that of OC-SVM by taking second order statistics of data into consideration.  Hence, single class MPM learns a decision boundary that generalizes well to the underlying data distribution.  Fig.~\ref{fig:other_methods} presents a high-level overview of different one-class classification methods. Though, these approaches are powerful tools in identifying the decision boundary for target data, their performance depends on the features used to represent the target class data.



Over the last five years, methods based on Deep Convolutional Neural Networks (DCNNs) have shown impressive performance improvements for object detection and recognition problems. Taking classification task as an example, the top-5 error rate of vision systems on the ImageNet dataset \cite{deng2009imagenet} has dropped from $\sim$ 25\% to 2.25\% in the last five years.  This has been made possible due to the availability of large annotated datasets, a better understanding of the non-linear mapping between input images and class labels as well as the affordability of Graphics Processing Units (GPUs).   These networks learn distinct features for a particular class against another using the cross entropy loss.  However, for one class problems training such networks in an end-to-end manner becomes difficult due to the absence of negative class data.

In recent years, several attempts have been made to counter the problem of training a neural network for one-class classification  \cite{sabokrou2018adversarially}, \cite{sabokrou2018deep}, \cite{chalapathy2018anomaly}, \cite{lawson2017finding}, \cite{ravanbakhsh2017abnormal}, \cite{zhou2017anomaly}, \cite{chalapathy2017robust}, \cite{perera2018learning}.
These approaches can be broadly classified in to two categories, generative approaches \cite{ravanbakhsh2017abnormal}, \cite{zhou2017anomaly}, \cite{chalapathy2017robust} and discriminative approaches \cite{chalapathy2018anomaly}, \cite{perera2018learning}.
Generative approaches use generative frameworks such as auto-encoders or Generative Adversarial Networks (GAN) \cite{goodfellow2014generative} for one-class classification.
For example, Ravanbakhsh et al. \cite{ravanbakhsh2017abnormal} and Sabokrou et al. \cite{sabokrou2018deep} proposed deep auto-encoder networks for event anomaly detection in surveillance videos.
However, in their approaches the focus is mainly on the image-level one-class classification.
Work by Lawson et al. \cite{lawson2017finding} developed a GAN-based approach for abnormality detection.  Sabokrou et al. \cite{sabokrou2018adversarially} extended that idea for detecting outliers from image data using an auto-encoder based generator with adversarial training.  In general, these generative models such as GANs are very difficult to train as compared to the discriminative classification networks \cite{GAN_training}.

Compared to the generative approaches, discriminative approaches for one-class classification have not been well explored in the literature. One such approach by Perera and Patel \cite{perera2018learning} utilize an external reference dataset as the negative class to train a deep network for one-class classification using a novel loss function.  In contrast to this method, we do not make use of any negative class data in our approach.   In another approach, Chalapathy et al. \cite{chalapathy2018anomaly} proposed a novel SVM inspired loss function to train a neural network for anomaly detection.  With some inspirations from other statistical approaches for one-class classification (i.e. taking origin as a reference to find the decision boundary), we propose a novel method called, One-Class CNN (OC-CNN),  to learn representations for one-class problems with CNNs trained end-to-end in a discriminative manner.    This paper makes the following contributions:
\begin{itemize}
	\item A new approach is proposed based on CNN for one class classification which is end-to-end trainable.
	\item Through experiments, we show that proposed approach outperforms other statistical and deep learning-based one class classification methods and generalizes well across a variety of one class applications.
\end{itemize}

\section{Proposed Approach}
\begin{figure}[t!]
	\includegraphics[height=4.5cm]{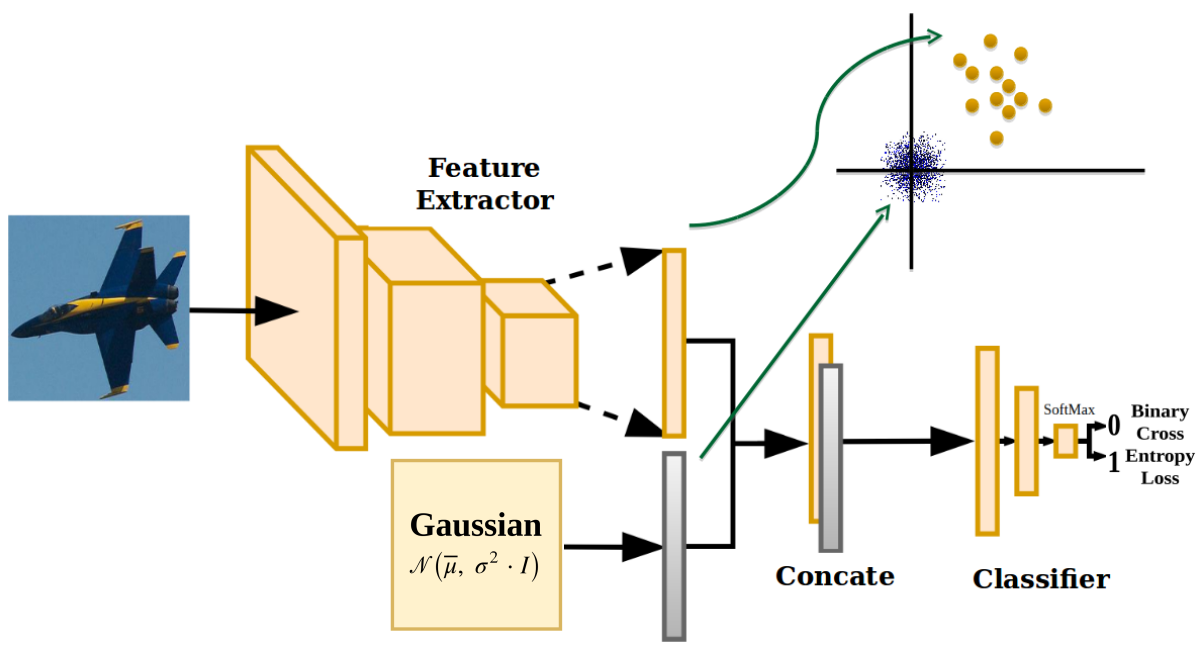}
\vskip -7.5pt \caption{Block diagram of the proposed approach. Here, $\bar{\mu}$ and $\sigma$ are mean and standard deviation parameters of a Gaussian, respectively and $\mathbf{I}$ is the identity matrix.}
	\label{fig:proposed_approach}
\end{figure}

Fig.~\ref{fig:proposed_approach} gives an overview of the proposed CNN-based approach for one-class classification.  The overall network consists of a feature extractor network and a classifier network.  The feature extractor network essentially embeds the input target class images into a feature space.  The extracted features are then appended with the pseudo-negative class data, generated from a zero centered Gaussian in the feature space. The appended features are then fed into a classification network which is characterized by a fully connected neural network.  The classification network assigns a confidence score for each feature representation.  The output of the classification network is either 1 or 0.  Here, 1 corresponds to the data sample belonging to the target class and 0 corresponds to the data sample belonging to the negative class. The entire nework is trained end-to-end using binary cross-entropy loss.
 
\subsection{Feature Extractor}
Any pre-trained CNN can be used as the feature extractor.  In this paper, we use the pre-trained AlexNet \cite{krizhevsky2012imagenet} and VGG16 \cite{simonyan2014very} networks by removing the softmax regression layers (i.e. the last layer) from their networks. During training, we freeze the convolution layers and only train the fully-connected layers.  Assuming that the extracted features are $D$-dimensional, the features are appended with the pseudo-negative data generated from a Gaussian, $ \mathcal{N}(\bar{\mu}, \sigma^2 \cdot \mathbf{I})$, where $\sigma$ and $\bar{\mu}$ are the parameters of the Gaussian and $\mathbf{I}$ is a $D\times D$ identity matrix. Here, $ \mathcal{N}(\bar{\mu}, \sigma^2 \cdot \mathbf{I})$ can be seen as generating $D$ independent one dimensional gaussian with $\sigma$ standard deviation.

\subsection{Classification Network}
Due to the appending of the pseudo-negative data with the original features, the classifer network observes the input in the batch size of 2.  A simple fully-connected layer followed by a softmax regression layer is used as the classifier network.  The dimension of the fully-connected layer is kept the same as the feature dimension.  The number of outputs from the softmax layer are set equal to two.

\subsection{Loss Function}
The following binary cross-entropy loss function is used to train the entire network
\begin{equation}\label{eq:cross_entropy_equation}
L_c \ = \ - \frac{1}{2K} \sum_{j=1}^{2K} \ (y \log(p) + (1-y) \log(1-p) ),
\end{equation}
where, $y \in \{0, 1\}$ indicates whether the classifier input corresponds to the feature extractor, (i.e. $y = 0$), or it is sampled from $ \mathcal{N}(\bar{\mu}, \sigma^2 \cdot \mathbf{I})$, (i.e. $y = 1$). Here, $p$ denotes the softmax probability of $y=0$.

The network is optimized using the Adam optimizer \cite{kingma2014adam} with learning rate of $10^{-4}$.  The input image batch size of 64 is used in our approach.  For all experiments, the parameters $\bar{\mu}$ and $\sigma$ are set equal to $\bar{0}$ and 0.01, respectively.  Instance normalization \cite{dumoulin2017learned} is used before the classifier network as it was found to be very useful in stabilizing the training procedure.

\begin{figure*}[!t]
	\centering
	\begin{subfigure}[t]{0.35\textwidth}
		\centering
		\includegraphics[height=1.0in]{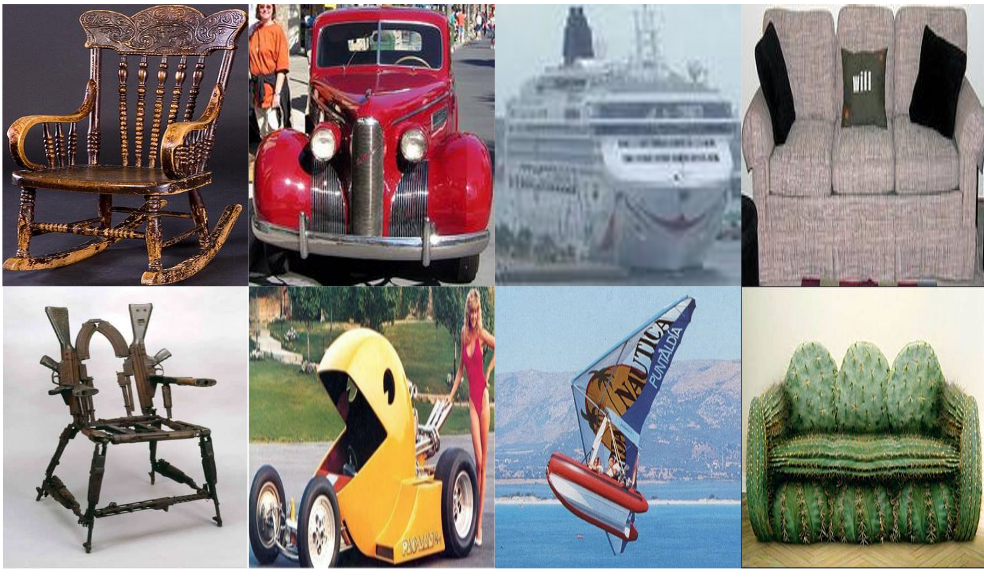}
		\vskip -5.0pt \caption{Abnormality-1001}
		\label{fig:sample_imagesa}
	\end{subfigure}%
	~ 
	\begin{subfigure}[t]{0.2\textwidth}
		\centering
		\includegraphics[height=1.0in]{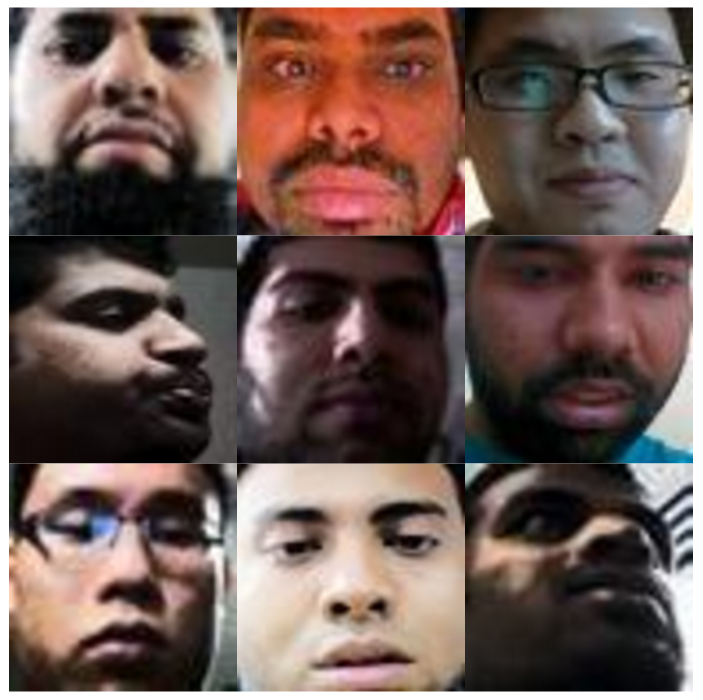}
		\vskip -5.0pt \caption{UMDAA-02 Face}
		\label{fig:sample_imagesb}
	\end{subfigure}
	~
	\begin{subfigure}[t]{0.2\textwidth}
		\centering
		\includegraphics[height=1.0in]{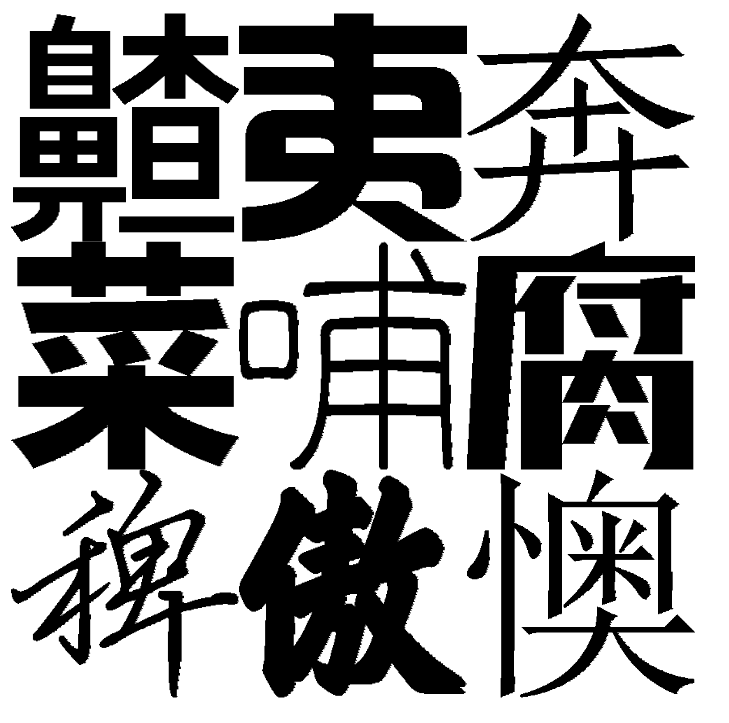}
		\vskip -5.0pt \caption{FounderType-200}
		\label{fig:sample_imagesc}
	\end{subfigure}    
\vskip -6.0pt	\caption{(a) Sample images from the Abnormality-1001 dataset \cite{saleh2013object}.   Normal and abnormal images are shown on the top row and the bottom row, respectively.  (b) Samples images from the UMDAA-02 Face Dataset \cite{mahbub2016active}.  (c)  Sample images from the FounderType-200 dataset \cite{liu2017incremental}. Different types of Chinese fonts.}
	\label{fig:sample_images}
\end{figure*}

\section{Experimental Results}\label{sec:experimental_details}

We evaluate the performance of the proposed approach on three different one-class classification problems - abnormality detection, face-based user authentication, and novelty detection.   Abnormality-1001 \cite{saleh2013object}, UMDAA-02 \cite{mahbub2016active} and FounderType-200 \cite{liu2017incremental} datasets are used to conduct experiments for the abnormality detection, user authentication and novelty detection problems. For all methods compared here, the data is aligned such that objects are at the center with minimal background.

The proposed approach is compared with following one-class classification methods:
\begin{itemize}
\item \textbf{OC-SVM:} One-Class Support Vector Machine is used as  formulated in \cite{scholkopf2000support}, trained using the AlexNet and VGG16 features.

\item \textbf{BSVM:} Binary SVM is used where the zero centered Gaussian noise is used as the negative data.  AlexNet and VGG16 features extracted from the target class data are used as the  positive class data.

\item \textbf{MPM:} MiniMax Probability Machines are used as  formulated in \cite{lanckriet2002minimax}. Since, the MPM algorithm involves computing covariance matrix from the data, Principal component analysis (PCA) is used to reduce the dimensionality of the features before computing the  covariance matrix.

\item \textbf{SVDD:} Support Vector Data Description is used as  formulated in \cite{tax2004support}, trained on the AlexNet and VGG16 features.

\item \textbf{OC-NN:} One-class neural network (OC-NN) is used as  formulated in \cite{chalapathy2018anomaly}. Here, for fair comparison, instead of using the feature extractor trained using an auto-encoder (as per \cite{chalapathy2018anomaly} methodology), AlexNet and VGG16 networks, the same as the proposed method, are used. As described in \cite{chalapathy2018anomaly}, we evaluate OC-NN using three different activation functions - linear, Sigmoid and ReLU.

\item \textbf{OC-CNN:} One-class CNN is the method proposed in this paper.

\item \textbf{OC-SVM$^+$:} OCSVM$^+$ is another method used in this paper where OC-SVM is utilized on top of the features extracted from the network trained using OC-CNN.  However, since it uses OC-SVM for classification, it is not end-to-end trainable.
\end{itemize}
\label{subsec:methods_compared}

\begin{table*}[!t]
\centering
\begin{tabular}{|c|c|c|c|c|c|c|c|c|c|}
\hline
\textbf{Dataset} & \textbf{OC-SVM} & \textbf{BSVM} & \textbf{MPM} & \textbf{SVDD} & \textbf{OC-NN-lin} & \textbf{OC-NN-sig} & \textbf{OC-NN-relu} & \textbf{OC-CNN} & \textbf{OC-SVM$^+$} \\ \hline
Abnormality-1001 & 0.6057 & 0.6126 & 0.5806 & 0.7873 & 0.8090 & 0.6391 & 0.7372 & \textit{0.8264} & \textbf{0.8334} \\ \hline
UMDAA-02 Face & 0.5746 & 0.5660 & 0.5418 & 0.6448 & 0.6173 & 0.6452 & 0.5943 & \textbf{0.7017} & \textit{0.6736} \\ \hline
FounderType-200 & 0.7124 & 0.7067 & 0.7085& 0.8998 & 0.8884 & 0.8696 & 0.8505 & \textit{0.9303} & \textbf{0.9350} \\ \hline
\end{tabular}
\vskip -5.0pt \caption{Comparison between the proposed and other methods using \textbf{AlexNet} as the base network. Results are mean of performance on all classes. Best and the second best performance are highlighted in bold fonts and italics, respectively.}
\label{table:auroc_Alex}
\end{table*}

\begin{table*}[!t]
\centering
\begin{tabular}{|c|c|c|c|c|c|c|c|c|c|}
\hline
\textbf{Dataset} & \textbf{OC-SVM} & \textbf{BSVM} & \textbf{MPM} & \textbf{SVDD} & \textbf{OC-NN-lin} & \textbf{OC-NN-sig} & \textbf{OC-NN-relu} & \textbf{OC-CNN} & \textbf{OC-SVM$^+$} \\ \hline
Abnormality-1001 & 0.6475 & 0.6418 & 0.5909 & 0.8031 & 0.7740 & 0.8373 & 0.5821 & \textit{0.8424} & \textbf{0.8460} \\ \hline
UMDAA-02 Face & 0.5829 & 0.5751 & 0.5473 & 0.6424 & 0.6193 & 0.6200 & 0.5788  & \textbf{0.7350} & \textit{0.7230} \\ \hline
FounderType-200 & 0.7490 & 0.7067 & 0.7444 & 0.8885 & 0.8986 & 0.8677 & 0.8506 & \textit{0.9290} & \textbf{0.9419}\\ \hline
\end{tabular}
\vskip -5.0pt \caption{Comparison between proposed and other methods using \textbf{VGG16} as the base network. Results are mean of performance on all classes. Best and the second best performance are highlighted in bold fonts and italics, respectively.}
\label{table:auroc_VGG}
\end{table*}

\subsection{Abnormality Detection}
Abnormality detection (also referred as anomaly detection or outlier rejection) deals with identifying instances that are dissimilar to the target class instances (i.e. abnormal instances).
Note that, the abnormal instances are not known a priori and only the normal instances are available during training.
Such problem can be addressed by one-class classification algorithms.
The Abnormality-1001 dataset \cite{saleh2013object} is widely used for visual abnormality detection.  This dataset consists of 1001 abnormal images belonging to six classes such as Chair, Car, Airplane, Boat, Sofa and Motorbike which have their respective normal classes in the PASCAL VOC dataset \cite{everingham2010pascal}.
Sample images from the Abnormality-1001 dataset are shown in Fig. \ref{fig:sample_images} (a).   Normal images obtained from the PASCAL VOC dataset are split into train and test sets such that the number of abnormal and normal images in test set are equal. Reported results are averaged for all six classes.

\subsection{User Active Authentication}
Active authentication refers to the problem of identifying the enrolled user based on his/her biometric data such as face, swipe patterns, and accelerometer patterns \cite{VMP_AA_SPM}.
The problem can be viewed as identifying the abnormal user behaviour to reject the unauthorized user.  The active authentication problem has been viewed as one-class classification problem \cite{antal2015evaluation}.
The UMDAA-02 dataset \cite{mahbub2016active} is widely used dataset for user active authentication on mobile devices.
The UMDAA-02 dataset has multiple modalities corresponding to each user such as face, accelerometer, gyroscope, touch gestures, etc.
Here, we only use the face data provided in this dataset since face is one of the most commonly used modality for authentication.
The face data consists of 33209 face images corresponding to 48 users.
Sample images corresponding to a few subjects from this dataset are shown in  Fig.~\ref{fig:sample_images}(b).
As can be seen from this figure, the images contains large variations in pose, illumination, appearance, and occlusions. For each class, train and test sets are created by maintaining 80/20 ratio.
Network is trained using the train set of a target user and tested on the test set of the target user against the rest of the user test set data.  This process is repeated for all the users and average results are reported.

\subsection{Novelty Detection}
The FounderType-200 dataset was introduced for the purpose of novelty detection by Liu et al. in \cite{liu2017incremental}.
The FounderType-200 dataset, contains 6763 images from 200 different types of fonts created by the company FounderType.
Fig.~\ref{fig:sample_images}(c) shows some sample images from this dataset.
For experiments, first 100 classes are used as the target classes and remaining 100 classes are used as the novel data.
The first 100 class data are split into train and test set having equal number of images.
For novel data, a novel set is created having 50 images from each of the novel classes.
For each class, train set from the known data is used for training the network and known class test set and novel set data are used for evaluation.
For example, class $i$ ($i \in \{1,2,..,100\}$) train set is used for training the network. The trained network is then evaluated with class $i$ test set tested against the novel set (containing data of class 101-200).
This is repeated for all classes $i$ where, $i \in \{1,2,..,100\}$ and average results are reported.

\section{Results and Discussion}
The performance is measured using the area under the receiver operating characteristic (ROC) curve (AUROC), most commonly used metric for one-class problems.   The results are tabulated in Table \ref{table:auroc_VGG} and Table \ref{table:auroc_Alex} corresponding to the VGG16 and AlexNet networks.  AlexNet and VGG16 pretrained features are used to compute the results for OC-SVM, BSVM, SVDD and MPM. The OC-NN results are computed using the linear, sigmoid and relu activations after training on the target class data.
The OC-CNN results are computed after training on the target class and for OC-SVM$^+$, an one-class SVM is trained on top of the features extracted from the trained AlexNet/VGG16, and AUROC is computed from the SVM classifier scores.

From the Tables \ref{table:auroc_Alex} and \ref{table:auroc_VGG}, it can be observed that either OC-CNN or OC-SVM$^+$ achieves the best performance on all three datasets.
MPM and OC-SVM achieve similar performances, while BSVM with Gaussian data as the negative class doesn't work as well.
With the BSVM baseline, we show that similar trick we used for proposed algorithm doesn't work well for statistical approaches like SVM.
Among the other one-class approaches, OC-NN with linear activation performs the best. However, OC-NN results are inconsistent.
For couple of experiments, SVDD was found to be working better than OC-NN.
The reason behind this inconsistent performance can be due to the differences in the evaluation protocol used for OC-NN in \cite{chalapathy2018anomaly} and this paper.
The ratio of the number of target class images to novel/abnormal class images in our evaluation protocol is much higher than the ratio used by Chalpathy et al. \cite{chalapathy2018anomaly}.
When the ratio is close to one, as is the case for Abnormality-1001 dataset, the OC-NN performs better than SVDD for both AlexNet and VGG16. However, when the ratio is increased (which is more realistic scenario), as is the case for UMDAA-02 and FounderType-200, the performance of OC-NN becomes inconsistent.
Whereas, using the proposed approach performs consistently well, providing  $\sim$4\%, $\sim$10\% and $\sim$5\% improvements over OC-NN for Abnormality-1001, UMDAA02- Face and FounderType-200 datasets, respectively.
Since, the proposed approach is built upon the traditional discriminative learning framework for deep neural networks, it is able to learn better features than OC-NN.

Also as expected, methods based on the VGG16 network work better than the methods based on the AlexNet network.
Apart from the FounderType-200 dataset where, OC-CNN with AlexNet works better than VGG16, for all methods VGG16 works better than AlexNet.
However, it should be noted that better OC-SVM$^+$ performance for VGG16 indicates that features learned with the proposed approach for VGG16 are better than AlexNet for FounderType-200.
Overall, VGG16 gives $\sim$2\% improvement over AlexNet.

Another interesting comparison is between OC-SVM and OC-SVM$^+$. OC-SVM uses features extracted from a pre-trained AlexNet/VGG16 network.
On the other hand, OC-SVM$^+$ uses features extracted from AlexNet/VGG16 network trained using the proposed approach. OC-SVM$^+$ performs $\sim$18\% and $\sim$17\% better than OC-SVM on average across all datasets for AlexNet and VGG16, respectively.
This result shows the ability of our approach to learn better representations. So, apart from being an end-to-end learnable standalone system, our approach can also be used to extract target class friendly features. Also, using sophisticated classifier has shown to improve the performance over OC-CNN (i.e., OC-SVM$^+$) in majority of cases.

\section{Conclusion}\label{sec:conclusion}
We proposed a new one-class classification method based on CNNs. A pseudo-negative Gaussian data was introduced in the feature space and the network was trained using a binary cross-entropy loss. 
Apart from being a standalone one-class classification system, the proposed method can also be viewed as good feature extractor for the target class data (i.e. OCSVM$^+$) as well. Furthermore, the consistent performance improvements over all the datasets related to authentication, abnormality and novelty detection showed the ability of our method to work well on a variety of one-class classification applications.  In this paper, experiments were performed over data with objects centrally aligned.  In the future, we will explore the possibility of developing an end-to-end deep one class method that does joint detection and classification.




\bibliographystyle{IEEETran}
\bibliography{refs}

\end{document}